\documentclass[10pt,twocolumn,letterpaper]{article}

\usepackage{iccv}
\usepackage{times}
\usepackage{epsfig}
\usepackage{graphicx}
\usepackage{amsmath}
\usepackage{amssymb}
\usepackage{float}
\usepackage{caption}

\usepackage{booktabs}

\usepackage{bm}
\usepackage{amsfonts}
\usepackage{algorithmic}
\usepackage{textcomp}
\usepackage{colortbl}

\usepackage{algorithm}
\usepackage{pifont} 
\usepackage{bbding} 
\usepackage{bbm}
\usepackage{indentfirst}

\usepackage{multirow}
\usepackage[T1]{fontenc}
\usepackage{color}

\usepackage[breaklinks=true,bookmarks=false]{hyperref}

\title{MoSAM: \underline{Mo}tion-Guided \underline{S}egment \underline{A}nything Model with Spatial-Temporal \underline{M}emory Selection}

\author{
    Qiushi Yang \quad
    Yuan Yao \quad
    Miaomiao Cui \quad
    Liefeng Bo \\[5pt]
    Institute for Intelligent Computing, Alibaba Group \\
    {\tt\small \{yangqiushi.yqs, ryan.yy, miaomiao.cmm, liefeng.bo\}@alibaba-inc.com} 
}

\begin{document}

\maketitle

\begin{abstract}
The recent Segment Anything Model 2 (SAM2) has demonstrated exceptional capabilities in interactive object segmentation for both images and videos. However, as a foundational model on interactive segmentation, SAM2 performs segmentation directly based on mask memory from the past six frames, leading to two significant challenges.
Firstly, during inference in videos, objects may disappear since SAM2 relies solely on memory without accounting for object motion information, which limits its long-range object tracking capabilities.
Secondly, its memory is constructed from fixed past frames, making it susceptible to challenges associated with object disappearance or occlusion, due to potentially inaccurate segmentation results in memory.
To address these problems, we present MoSAM, incorporating two key strategies to integrate object motion cues into the model and establish more reliable feature memory.
Firstly, we propose Motion-Guided Prompting (MGP), which represents the object motion in both sparse and dense manners, then injects them into SAM2 through a set of motion-guided prompts. MGP enables the model to adjust its focus towards the direction of motion, thereby enhancing the object tracking capabilities.
Furthermore, acknowledging that past segmentation results may be inaccurate, we devise a Spatial-Temporal Memory Selection (ST-MS) mechanism that dynamically identifies frames likely to contain accurate segmentation in both pixel- and frame-level. By eliminating potentially inaccurate mask predictions from memory, we can leverage more reliable memory features to exploit similar regions for improving segmentation results.
Extensive experiments on various benchmarks of video object segmentation and video instance segmentation demonstrate that our MoSAM achieves state-of-the-art results compared to other competitors.

\end{abstract}

\vspace{-0.4cm}
\section{Introduction}
\vspace{-0.2cm}

\label{sec:intro}

Segment Anything Model 2 (SAM2)~\cite{sam2} has demonstrated remarkable advances in both image and video object segmentation through interactive manner. Trained on the extensive Segment Anything Video dataset, which comprises 35.5M masks across 50.9K videos, SAM2 exhibits strong generalization capabilities across various downstream applications~\cite{medseg1,xu2024embodiedsam,mobile_sam,grounded-sam2,Semantic-SAM}, including medical video segmentation, 3D point cloud segmentation, robotics learning, etc., delivering accurate spatial-temporal localization and segmentation results.

As a representative foundation model on interactive visual segmentation, SAM2 performs frame-by-frame object segmentation guided by first-frame prompts and maintains a memory bank of past frames to facilitate subsequent frame segmentation through similar region activation. Despite its efficiency, SAM2 has two notable limitations in its design. Firstly, during frame sequence segmentation, SAM2 primarily relies on features from past frames for guidance, neglecting object motion information. This oversight often leads to tracking failures, particularly in frames with object occlusions or disappearance. Secondly, SAM2 indiscriminately stores frame features of fixed past frames in its memory bank. In cases involving object occlusion or disappearance, these stored features may lack meaningful object information, resulting in the accumulation of noisy features that offer limited or misleading guidance for segmentation. Several recent concurrent works~\cite{sam2long,yang2024samurai} propose constructing more reliable memory banks by selecting frames based on frame-level confidence scores and motion-guided cues. However, they ignore two important aspects: the reliability of regions within individual frames and the explicit incorporation of motion information for model prompting.

To address the aforementioned challenges, we propose a comprehensive framework that emphasizes two key aspects: motion-aware tracking and selective memory maintenance. To address these issues, we recognize that incorporating motion information is crucial for predicting object locations and maintaining temporal consistency. This insight motivates us to explore both sparse and dense motion representations, which can provide predictive cues towards object movement patterns and potential locations in subsequent frames. Such motion-aware design enables the model to better handle cases of object occlusion and temporary object disappearance by anticipating their localizations.
For the memory bank quality issue, we realize that a more discriminative approach to memory management is essential. Rather than storing features indiscriminately, we propose evaluating the reliability of frame features at both temporal and spatial levels. This dual-perspective assessment allows us to maintain a high-quality memory bank by selecting the most informative frames and focusing on reliable regions within those frames, thereby reducing the negative impact of noisy or irrelevant features.

In this work, we introduce MoSAM, which comprises two key strategies to address the issues of object disappearance and memory reliability. Specifically, to provide object motion cues for segmentation across frames, we design Motion-Guided Prompting (MGP), which extracts object motion representations through both sparse-level point movement and dense-level optical flow. These representations are then utilized to estimate subsequent localizations based on the mask regions of previous frames. This spatial forecasting is employed to warp the current prompt, thereby updating it for segmentation in the next frame.
Furthermore, to ensure a reliable and effective memory bank, we propose a Spatial-Temporal Memory Selection (ST-MS) mechanism. ST-MS begins by selecting more reliable frame features to serve as the memory bank, ranking the IoU scores from a few preceding frames. Afterwards, it filters out less confident pixel predictions within each memory feature. This mechanism allows MoSAM to extract features from relatively reliable frames and pixels in both temporal and spatial dimensions, ultimately boosting video segmentation.
Extensive experiments on various video object segmentation and video instance segmentation benchmarks demonstrate that MoSAM outperforms existing methods, achieving state-of-the-art performance.
In summary, our main contributions are fourfold:
\vspace{-2pt}
\begin{itemize}
\item We present MoSAM, a unified framework that synergistically integrates MGP and ST-MS to enhance motion-aware segmentation with reliable memory management.
\item To provide motion cues for the model, facilitating superior object tracking and segmentation, MGP captures the motion representation in both sparse and dense manners and then forecasts the subsequence object localization as future prompts.
\item Considering that the SAM2 memory bank may contain unreliable frame features without objects, ST-MS is designed to adaptively pick up more reliable frame features to update the memory bank by using confidence from both temporal and spatial levels.
\item Comprehensive experiments on various video object and video instance segmentation benchmarks verify that our MoSAM exhibits state-of-the-art performance. In particular, MoSAM boosts the average results over the baseline of 4.4\%, 3.0\%, 1.9\% in three datasets, respectively.
\vspace{-0.2cm}
\end{itemize}

\begin{figure*}
\begin{center}
\setlength{\belowcaptionskip}{-0.1cm}
\includegraphics[width=0.98\linewidth]{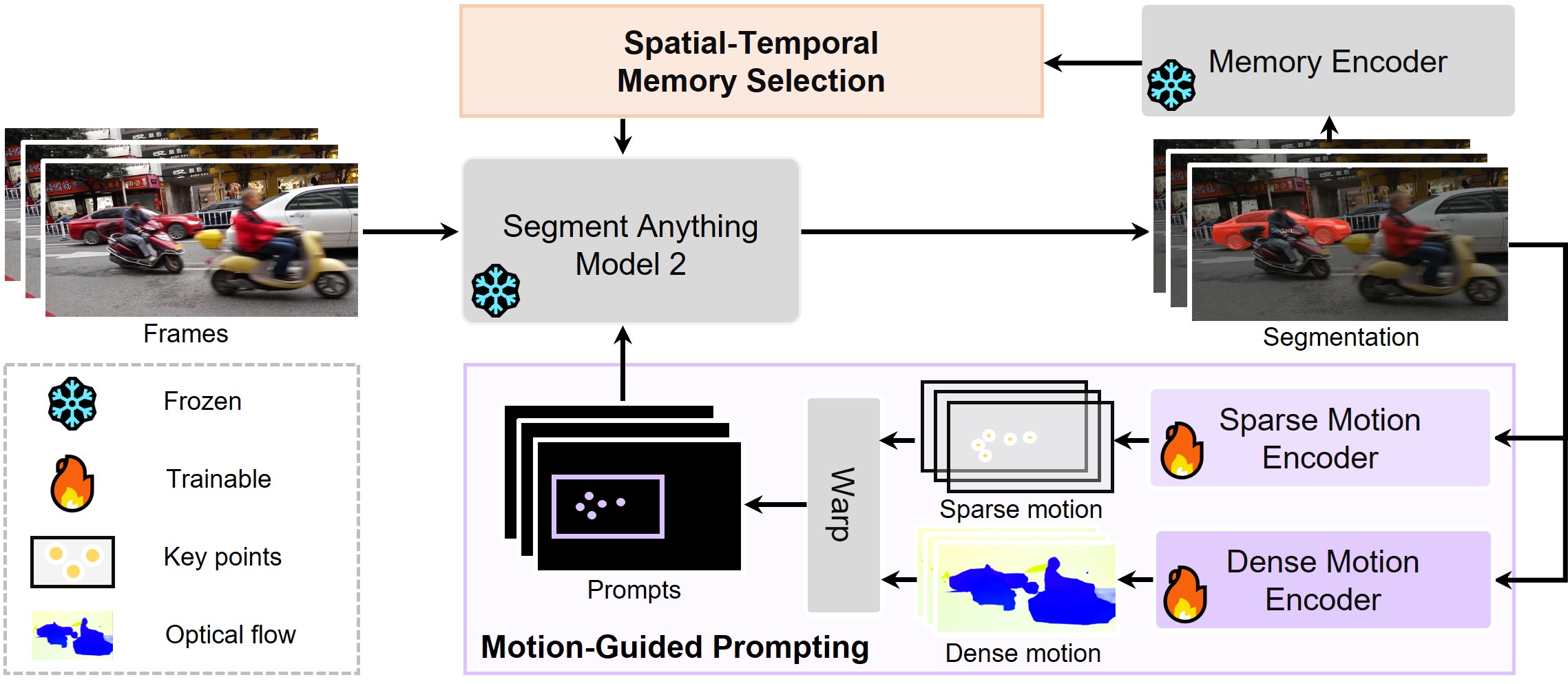}
\caption{
Overview of the proposed MoSAM framework. It consists of a motion-guided prompting (MGP) mechanism to inject object motion cues into the model for superior object tracking, and a spatial-temporal memory selection (ST-MS) strategy, dynamically updating feature memory to maintain a reliable and effective memory bank.
}
\label{fig:network}
\end{center}
\end{figure*}

\section{Related Work}
\label{sec:related}

\vspace{-2pt}
\subsection{Video Object Segmentation}
\vspace{-2pt}
Video Object Segmentation (VOS)~\cite{yang2019video} focuses on identifying and localizing objects within a video sequence, based on a single annotated object frame, which is referred to as semi-supervised VOS.
Early VOS approaches~\cite{one_shot,monet,onlinevos,maskrnn,premvos} employ online inference to adapt pre-trained segmentation models for recognizing specified objects. However, this online segmentation manner is time-consuming, significantly slowing down inference.
To solve this problem, propagation-based methods~\cite{sstvos,fastvos,vosflow} leverage temporal correlations between adjacent frames through offline shift attention learning, allowing for the propagation of object masks from previous frames to current ones. While these methods show promising results, they are prone to error accumulation stemming from inaccurate predictions.

In pursuit of rapid and efficient inference, matching-based methods~\cite{pml,videomatch,lwl,cfbi,SimVOS,joint} identify targets by comparing the object template to the test image, predicting object masks based on matching features. For example, VideoMatch~\cite{videomatch} performs pixel matching among adjacent frames to generate mask predictions.
Some other approaches~\cite{STM,aot,deaot,XMem} enhance the matching-based VOS methods by incorporating object memory to enrich object representations and by developing advanced matching strategies to produce more comprehensive correlated features. For instance, JointFormer~\cite{joint} proposes to adopt Transformer blocks to jointly model features and patch correspondences for object representation and achieves decent segmentation results. 
Despite effectiveness, these approaches mainly focus on specific datasets and categories and cannot be generalized to open-world video object segmentation scenarios.

\vspace{-2pt}
\subsection{Segment Anything Model}
\vspace{-2pt}
The Segment Anything Model (SAM)~\cite{sam} has been pre-trained on massive dense annotated datasets, established as a robust benchmark for general visual segmentation tasks. Numerous follow-up studies have expanded upon SAM, creating various adaptations for specific downstream applications~\cite{Gaussian-Grouping,SEEM,Semantic-SAM,ma2024segment,xu2024embodiedsam}.
For instance, since SAM lacks the capability for semantic prediction, some researches~\cite{Semantic-SAM,SEEM} incorporate category labels to fine-tune the model, enabling it to perform semantic segmentation. SEEM~\cite{SEEM} enhances SAM's functionality by training it with labeled segmentation data, employing a bipartite matching constraint to enable semantic prediction. Similarly, Semantic-SAM~\cite{Semantic-SAM} introduces a multi-choice learning framework through multi-task training across different datasets, which allows the model to segment at various levels of detail while also predicting semantic labels.
Moreover, other studies~\cite{ma2024segment,xu2024embodiedsam,Gaussian-Grouping} adapts SAM for specialized domains to tackle specific application challenges. For example, MedSAM~\cite{ma2024segment} fine-tunes SAM using a comprehensive dataset of medical images across various modalities and cancer types, resulting in improved performance in disease segmentation tasks. Gaussian-Grouping~\cite{Gaussian-Grouping} utilizes SAM as a supervisory model to concurrently train for object reconstruction and segmentation within open-world 3D scenes, facilitating high-quality 3D editing.

More recently, SAM2~\cite{sam2} extend the SAM to video segmentation, which introduces a feature memory bank as the object feature template for segmentation in a online frame-by-frame manner. Some concurrent studies~\cite{sam2long,yang2024samurai} revise SAM2 by enhancing memory bank for better object tracking. However, these works focus merely on the temporal frame-level memory selection, neglecting the spatial pixel-level filtering issue. Instead, we consider comprehensive spatial and temporal level memory selection, and aim to explicitly provide motion information to the SAM2 model for superior segmentation capability.


\vspace{-2pt}
\section{Methodology}
\label{sec:method}

\vspace{-2pt}
\subsection{Preliminaries: SAM2}
SAM2~\cite{sam2} conducts video segmentation in an online frame-by-frame manner. Building on SAM~\cite{sam}, SAM2 creates a memory bank that stores past mask predictions along with their corresponding features. This memory bank is used to activate the object regions in current frame $v_t$, where $t$ is the $t$-th frame, through a cross-attention module, which is updated by $N$ nearest frames with the first prompted frame of video $V$ in a first-in-first-out manner.

The output mask prediction in SAM2 is generated from the output mask token. Meanwhile, SAM2 predicts the Intersection over Union (IoU) score $S_{\text{IoU}}$ for each mask output, and an occlusion score $S_{\text{Occ}}$ for each frame. The $S_{\text{IoU}}$ score reflects the confidence of the prediction regarding the overlap between the output mask and the true object regoin, while $S_{\text{Occ}}$ assesses the confidence of occlusion conditions. A value of $S_{\text{Occ}} \geq 0$ means the presence of the object.

\vspace{-2pt}
\subsection{Motion-Guided Prompting (MGP)}
\vspace{-2pt}
Since SAM2 overlooks the object motion cues and cannot tracks the object well, especially in object occlusion and disappearance conditions. To address this problem, we propose to represent the motion information according to the past few mask predictions, and then predict the future object localization for the next frame, which is used to warp the prompt as a type of future localization prompt to be injected into the model to perform following segmentation with the motion cues. We aim to obtain both geometric- and region-aware prompts via sparse and dense motion representation, which are described in detail below.

\vspace{-2pt}
\subsubsection{Sparse Motion Modeling}
Given the $t$-th input frame $v_t$ within a video $V$, SAM2 produces the final feature $f_t$ and the corresponding mask prediction $m_t$.
To obtain the robust object motion representation for mask prediction and accommodate potential abrupt changes in future motion, we propose the sparse motion modeling. 
Leveraging the current features $f_t$ and the mask prediction $m_t$,
we first extract multiple geometric key points from within the object area to represent its geometric information $p_t=\phi_{S}(f_t, m_t)$, where $\phi_{S}$ means the sparse motion encoder. This includes the geometric centroid of the object region and points located at half the distance from the centroid to the vertical edges in the up, down, left, and right directions. Similarly, we extract geometric key points for the same object from the past prediction using the same approach $p_{t-1}=\phi_{S}(f_{t-\Delta{t}},m_{t-\Delta{t}})$.
Acquiring these sparse geometric key points for the object in both frames, we compute the movement information for each corresponding key point between the two frames, which includes both the direction and distance of movement. Using the movement states of the past few key points, we then predict the spatial positions of the key points for the next frame through linear interpolation. We incorporate these estimated future key points as a sparse representation of object motion to infuse into the model in the form of positive point prompts. This informs the model of the object's potential future locations, facilitating more accurate object segmentation.

\vspace{-2pt}
\subsubsection{Dense Motion Modeling}
To yield the global localization of the object, we further formulate the motion representations for the objects according to the pixel-level movement. We capture the dense pixel motion cues by yielding the optical flow between the current and previous object feature guided by the mask: $g_d=m_t \cdot \phi_{D}(f_t,f_{t-\Delta{t}})$, where $m_t$ is the current mask prediction and $\phi_{D}$ refers to the dense motion encoder, here we employ an optical flow estimation network. This dense motion representation exhibits the movement intensity for each pixel in both horizon and vertical directions among the current and the past frames. 
Upon the object motion information derived from the current and previous frames, we employ linear interpolation to forecast the future motion state and object position in the next frame. Subsequently, we warp the current mask prediction accordingly to estimate the probable location of the object in the next frame. Utilizing the predicted object position region, we extract the corresponding bounding box as a box prompt, which is then fed into the model to serve as a positional cue for segmentation in the subsequent frame. By modeling the dense motion cues, we can estimate the object localization and inject it into the model via the prompt form, thereby enhancing the model's capability to perceive changes in object position and accurately track the object.

In MGP, the sparse motion cues offer the geometric key point prompts to maintain the robustness of the motion representation, and the dense motion provides global spatial region information as the box prompt. 
These two object motion representation strategies can complement each other to provide the model with object motion information. By roughly approximating the future spatial positions of the objects, they enhance the model's ability to accurately track and segment the objects.

\vspace{-2pt}
\subsection{Spatial-Temporal Memory Selection (ST-MS)}
\begin{figure}[t]
\begin{center}
\setlength{\belowcaptionskip}{-0.1cm}
\includegraphics[width=\linewidth]{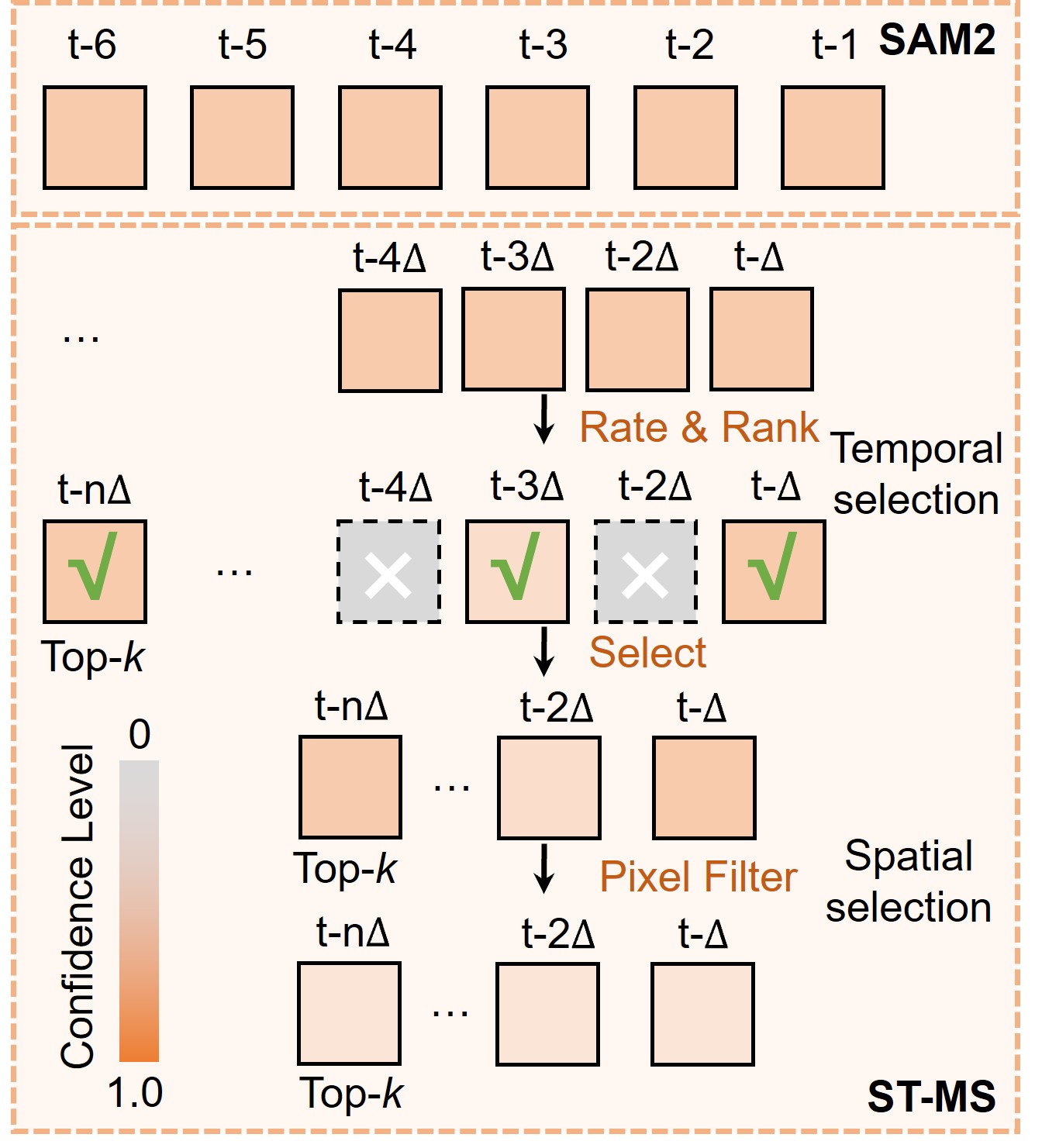}
\caption{
Illustration of the proposed spatial-temporal memory selection (ST-MS) strategy, which can pick up relative confident frame features in the temporal level and the reliable pixels of each frame in the spatial level to update the memory bank. 
} 
\label{fig:st-ms}
\end{center}
\end{figure}

\vspace{-2pt}
In the segmentation process for a video, SAM2 maintains a memory bank that contains past segmentation features and adopts cross-attention with the features in the memory bank to locate the object in the current frame. Nevertheless, SAM2 merely updates the memory bank with features from the first frame and a fixed set of the past six frames. This can lead to the storage of incorrectly predicted features or features from frames where the object is completely absent, especially in cases of object occlusion or temporary disappearance. Consequently, this results in a continuous accumulation of errors in the memory bank, adversely affecting the segmentation performance in subsequent frames.
To address this issue, as illustrated in Figure~\ref{fig:st-ms}, we propose a Spatial-Temporal Memory Selection (ST-MS) mechanism, which selectively filters each frame based on prediction confidence and further filters each pixel based on segmentation confidence, ensuring that only the most reliable frame pixels are stored in the memory bank as target features.

\begin{table*}[t]
\renewcommand\arraystretch{1}
\setlength{\tabcolsep}{5.8pt}
\centering
\footnotesize
\caption{Performance comparison between the baseline SAM2 and MoSAM across various model sizes, including Tiny (-T), Small (-S), Base (-B+) and Large (-L). Result gains over the baseline by our method are in \textcolor{red}{red}.}
\label{baseline-comparison}
\begin{tabular}{lccc|ccc|ccc}
\toprule
 \multicolumn{1}{l}{\multirow{2}{*}{Methods}} & \multicolumn{3}{c|}{LVOS v1} & \multicolumn{3}{c|}{SA-V val} &  \multicolumn{3}{c}{SA-V test}  \\
 &  $\mathcal{J}\&\mathcal{F}$ & $\mathcal{J}$  & $\mathcal{F}$ & $\mathcal{J}\&\mathcal{F}$ & $\mathcal{J}$  & $\mathcal{F}$ &  $\mathcal{J}\&\mathcal{F}$ & $\mathcal{J}$  & $\mathcal{F}$ \\\midrule
SAM2-T~\cite{sam2}   & 77.5 & 73.0 & 82.1   & 75.1 & 71.6 & 78.6 & 76.3 & 72.7 & 79.8 \\
MoSAM-T & 81.1 (\textcolor{red}{+3.6}) & {75.5} (\textcolor{red}{+2.5}) & 86.6 (\textcolor{red}{+4.5}) & 79.7 (\textcolor{red}{+4.6}) & 75.9 (\textcolor{red}{+4.3}) &   83.5 (\textcolor{red}{+4.9})  & 79.8 (\textcolor{red}{+3.5}) & 76.0 (\textcolor{red}{+3.3}) &  86.3 (\textcolor{red}{+6.5}) \\ \hline

SAM2-S~\cite{sam2}   & 77.3 & 72.3 & 82.2    & 76.9 & 73.5 & 80.3 & 76.9 & 73.3 & 80.5 \\
MoSAM-S & {81.9} (\textcolor{red}{+4.6}) & {76.2} (\textcolor{red}{+3.9}) & {87.7} (\textcolor{red}{+5.5})         & {79.6} (\textcolor{red}{+2.7}) & {75.8} (\textcolor{red}{+2.3}) & {83.3} (\textcolor{red}{+3.0}) & {80.5} (\textcolor{red}{+3.6}) & {76.3} (\textcolor{red}{+3.0}) & {84.7} (\textcolor{red}{+4.2}) \\  \hline

SAM2-B+~\cite{sam2}   & 77.7 & 73.1 & 82.4    & 78.0 & 74.6 & 81.5 & 77.7 & 74.2 & 81.2 \\
MoSAM-B+ & {82.1} (\textcolor{red}{+4.4}) & {76.9} (\textcolor{red}{+3.8}) & {87.4} (\textcolor{red}{+5.0}) & {80.1} (\textcolor{red}{+2.1}) & {76.4} (\textcolor{red}{+1.8}) & {83.9} (\textcolor{red}{+2.4}) & {80.2} (\textcolor{red}{+1.9}) & {76.9} (\textcolor{red}{+2.7}) & {83.6} (\textcolor{red}{+2.4}) \\  \hline

SAM2-L~\cite{sam2}   & 80.2 & 75.4 & 84.9   & 78.6 & 75.1 & 82.0 & 79.6 & 76.1 & 83.2 \\
MoSAM-L & {84.6} (\textcolor{red}{+4.4}) & {79.3} (\textcolor{red}{+3.9}) & {89.9} (\textcolor{red}{+5.0}) & {81.6} (\textcolor{red}{+3.0}) & {77.9} (\textcolor{red}{+2.8}) & {85.4} (\textcolor{red}{+3.4}) & {81.5} (\textcolor{red}{+1.9}) & {77.7} (\textcolor{red}{+1.6}) & {85.3} (\textcolor{red}{+1.9}) \\ 

\bottomrule
\end{tabular}
\end{table*}

\vspace{-2pt}
\subsubsection{Temporal Memory Selection}
\vspace{-2pt}
Considering that adjacent frames may contain much redundant information, we sample past frames at regular intervals $\Delta t$. Based on the IoU score $S_\text{IoU}$ and occlusion score $S_\text{Occ}$ obtained from SAM2 for each frame, we first filter out a certain number of frames from the past video frames within a defined range that do not exhibit occlusion and have relatively high prediction confidence $v_t, v_t(S_\text{IoU})>\text{thr}_\text{IoU}, v_t(S_\text{Occ})>\text{thr}_\text{Occ}$. These selected frames are stored in the memory. Then, we sort the remaining frames based on the total score derived from the sum of the IoU score and occlusion score, selecting the highest-scoring frames to store in the memory.

In this way, we select relatively reliable target predictions at the frame level based on the model's predicted confidence. The filtered frames are more likely to contain the target area, enhancing the quality and reliability of the memory bank, thereby assisting the model in making better segmentation predictions.

\vspace{-2pt}
\subsubsection{Spatial Memory Selection}
\vspace{-2pt}
Given the candidate frames filtered at the frame level, we further apply spatial filtering to the mask predictions within each frame, retaining more confident predicted areas as the foreground and discarding the less confident areas as the background. We derive a probability map for each frame's mask prediction from SAM2, and apply a threshold to filter the confidence of each pixel predicted as the foreground. Pixels with confidence above the threshold are retained as object region.
This mechanism refines the initial mask predictions, yielding more accurate masks that serve as features in the memory bank. It alleviates the subsequent impact of erroneous object position predictions and enhances the segmentation accuracy in later frames.

Through the selection and filtering of candidate memory bank features at both the temporal frame-level and spatial pixel-level, we can obtain more reliable object features for the memory bank. This provides a more accurate feature template for the object segmentation in subsequent frames.


\begin{table*}[t]
\renewcommand\arraystretch{1}
\setlength{\tabcolsep}{6.5pt}
\centering
\footnotesize
\caption{Comparison of the MoSAM with state-of-the-arts on video object segmentation task. All SAM-based models adopt the large variant architecture of the latest SAM2.1 version. \textbf{Bold} suggests the best results.}
\vspace{2pt}
\label{vos-comparison}
\begin{tabular}{lccc|ccccc|ccc|ccc}
\toprule
 \multicolumn{1}{l}{\multirow{2}{*}{Methods}} & \multicolumn{3}{c|}{LVOS v1} & \multicolumn{5}{c|}{LVOS v2} & \multicolumn{3}{c|}{SA-V val} &  \multicolumn{3}{c}{SA-V test}  \\  
 &  $\mathcal{J}\&\mathcal{F}$ & $\mathcal{J}$  & $\mathcal{F}$  &  $\mathcal{J}\&\mathcal{F}$ & $\mathcal{J}_s$  & $\mathcal{F}_s$ & $\mathcal{J}_u$ & $\mathcal{F}_u$ &  $\mathcal{J}\&\mathcal{F}$ & $\mathcal{J}$  & $\mathcal{F}$ &  $\mathcal{J}\&\mathcal{F}$ & $\mathcal{J}$  & $\mathcal{F}$ \\\midrule
LWL~\cite{lwl}                      & 56.4 & 51.8 &   60.9    & 60.6 & 58.0 & 64.3 & 57.2 & 62.9  & - &  -  & - & - &   -    & -  \\ 
CFBI~\cite{cfbi}                    & 51.5 & 46.2 &   56.7    & 55.0 & 52.9 &  59.2  & 51.7 & 56.2 &   -    & - & - &  - &  - &  - \\ 
STCN~\cite{STCN}                    & 48.9 & 43.9 &   54.0    & 60.6 & 57.2 &  64.0  & 57.5 & 63.8 &   61.0    & 57.4 & 64.5 &  62.5 &  59.0 &  66.0 \\ 
RDE~\cite{RDE}                      & 53.7 & 48.3 &   59.2    & 62.2 & 56.7 &  64.1  & 60.8 & 67.2 & 51.8 & 48.4 & 55.2    & 53.9 & 50.5 &  57.3 \\ 
SwinB-DeAOT-L~\cite{SwinB-DeAOT-L}  & - & - & - & 63.9 & 61.5 &  69.0  & 58.4 & 66.6 &  61.4    & 56.6 & 66.2 &  61.8 & 57.2 & 66.3 \\  
XMem~\cite{XMem}                    & 52.9 & 48.1 &   57.7    & 64.5 & 62.6 &  69.1  & 60.6 & 65.6 &   60.1    & 56.3 & 63.9 &  62.3 & 58.9 & 65.8 \\  
DEVA~\cite{DEVA}             & 55.9 & 51.1 & 60.7 & - & - &  -  & - & -   & 55.4 & 51.5 & 59.2 & 56.2 & 52.4 & 60.1 \\
Cutie-base~\cite{Cutie-base} & 66.0 & 61.3 & 70.6 & - & - &  -  & - & -   & 60.7 & 57.7 & 63.7 & 62.7 & 59.7 & 65.7 \\
\midrule
SAM2~\cite{sam2}   & 80.2 & 75.4 &  84.9  & 84.1 & 80.7 &  87.4 & 80.6 &  87.7 & 78.6 & 75.1 & 82.0 & 79.6 & 76.1 & 83.2 \\
SAM2Long~\cite{sam2long} & 83.4 & {78.4} & 88.5  & \textbf{85.9} & \textbf{81.7} & \textbf{88.6} &  \textbf{83.0} & {90.5} & 81.1 & 77.5 & 84.7 & 81.2 & 77.6 & 84.9 \\
SAMURAI~\cite{yang2024samurai} & 81.8 & {76.7} & 86.9  & 84.2 & {78.6} & {88.5} &  79.4 &  {90.5}  & 79.8 & 75.9 &   83.6    & 80.0 & 76.2 &  83.9 \\
MoSAM    & \textbf{84.6} & \textbf{79.3} & \textbf{89.9} & {85.7} & {78.4} & {88.2} &  {82.5} &  \textbf{93.5} & \textbf{81.6} & \textbf{77.9} & \textbf{85.4} & \textbf{81.5} & \textbf{77.7} & \textbf{85.3} \\ 
\bottomrule
\end{tabular}
\end{table*}

\vspace{-2pt}
\section{Experiments}
\vspace{-2pt}
\label{sec:exp}

\subsection{Datasets and Details}
\vspace{-2pt}
To train and assess our proposed method, we adopt the DAVIS2017 as the training set and use three public video object segmentation (VOS) datasets and one video instance segmentation (VIS) set to evaluate all methods.

\noindent\textbf{DAVIS2017} is a benchmark for VOS, comprising 150 videos with detailed mask annotations: 60, 30, 60 for training, validation, and testing. We utilize both the training and validation subsets to optimize our framework.

\noindent\textbf{LVOS-v1} serves as a long-term VOS benchmark in realistic settings, featuring 720 video clips with 296,401 frames and 407,945 annotations, averaging over 60 seconds in duration. This dataset introduces complexities like long-term object reappearance and temporally similar objects.

\noindent\textbf{LVOS-v2} builds on LVOS-v1, offering 420, 140, and 160 videos for training, validation, and testing, respectively. It includes 44 categories, with 12 held back to assess the generalization capabilities of VOS models.

\noindent\textbf{SA-V} is a large-scale VOS dataset, featuring 50.9K video clips and 642.6K masklets, totaling 35.5 million annotated masks. It presents challenges such as small, occluded, and reappearing objects. The validation set includes 293 masklets across 155 videos, while the testing set contains 278 masklets across 150 videos.

\noindent\textbf{LV-VIS} is a comprehensive VIS dataset with 4,828 real-world videos across 1,196 categories. The split includes 3,083 for training, 837 for validation, and 908 for testing. The classes are divided into 641 base categories and 555 novel categories.

\noindent\textbf{Evaluation Metrics} 
For VOS, we adopt $J$ (region similarity), $F$ (contour accuracy), and the combined $J\&F$ scores, along with additional metrics $J_s, F_s, J_u, F_u$ for LVOS-v2, evaluated in a semi-supervised setting with a mask prompt provided for the first frame. For VIS, we calculate the mean Average Precision (mAP) across all categories, detailing mAP$_b$ for base categories and mAP$_n$ for novel categories.

\noindent\textbf{Implementation Details}
During training, we fix the parameters of the pre-trained SAM2 model and train the sparse and dense motion encoder on the DAVIS dataset. We evaluate various SAM2 backbones of different sizes to assess the framework's robustness. The learning rate is set to 0.001, with training conducted over 30 epochs using a decay schedule. We utilize 8 NVIDIA A100 GPUs with a total batch size of 32. Following SAM2, in ST-MS, the memory bank length is set to seven, and the thresholds $\tau_\text{IoU}$ and $\tau_\text{Occ}$ are 0.7 and 0.0, respectively.

\vspace{-2pt}
\subsection{Comparison with SAM2}
\vspace{-2pt}
To assess the effectiveness of our MoSAM, which is built upon the baseline SAM2, we conduct comprehensive comparisons of VOS performance across various model architectures, including Tiny (-T), Small (-S), Base (-B+), and Large (-L) variants, on LVOS-v1, SA-V validation, and SA-V test benchmarks. As demonstrated in Table~\ref{baseline-comparison}, MoSAM consistently achieves superior performance over SAM2 across all model scales and evaluation datasets. Notably, our largest model variant (MoSAM-L) demonstrates substantial improvements in LVOS-v1, with gains of 4.4\%, 3.9\%, and 5.0\% in $\mathcal{J}\&\mathcal{F}$, $\mathcal{J}$, and $\mathcal{F}$ metrics, respectively. Similar performance gains are observed in SA-V validation set (3.0\%, 2.8\%, 3.4\%) and SA-V test set (1.9\%, 1.6\%, 1.9\%). These consistent improvements across different architectural configurations strongly indicate that our framework, incorporating motion-prompting strategy and spatial-temporal memory selection mechanism, significantly enhances the model's segmentation capabilities.

\vspace{-2pt}
\subsection{Comparison with Existing Methods on VOS}
\vspace{-2pt}
To comprehensively evaluate the capabilities of the proposed MoSAM in multiple VOS benchmarks, we compare it against various previous approaches, including traditional close-set video segmentation algorithms~\cite{lwl,cfbi,STCN,RDE,SwinB-DeAOT-L,XMem,DEVA,Cutie-base} and the latest SAM-based methods~\cite{sam2,sam2long,yang2024samurai}. As suggested in Table~\ref{vos-comparison}, MoSAM achieves scores of 84.6\%, 79.3\%, and 89.9\% for $\mathcal{J}\&\mathcal{F}$, $\mathcal{J}$, and $\mathcal{F}$ on LVOS-v1 and 85.7\%, 78.4\%, 88.2\%, 82.5\% and 93.5\% for $\mathcal{J}\&\mathcal{F}$, $\mathcal{J}_s$, $\mathcal{F}_s$, $\mathcal{J}_u$, and $\mathcal{F}_u$ on LVOS-v2. Moreover, we obtain scores of 81.6\%, 77.9\%, 85.4\% on the SA-V validation set, and 81.5\%, 77.7\%, 85.3\% on the SA-V test set. We attain state-of-the-art results across most datasets and metrics. These performance advantages demonstrate the effectiveness of our proposed framework, which utilizes sparse and dense motion representation to offer future movement cues and makes the feature memory reliable via spatial-temporal memory selection. MoSAM significantly impacts videos containing various objects and complex scenarios, e.g., occlusions and disappearance. On LVOS-v2, a recent concurrent work SAM2Long~\cite{sam2long} slightly outperforms our method, likely due to its use of a more complex ensemble and selection strategy with additional memory paths, leading to more computational overhead. In contrast, our method achieves commendable results using a simpler strategy.

\begin{figure*}[t]
\begin{center}
\includegraphics[width=\linewidth]{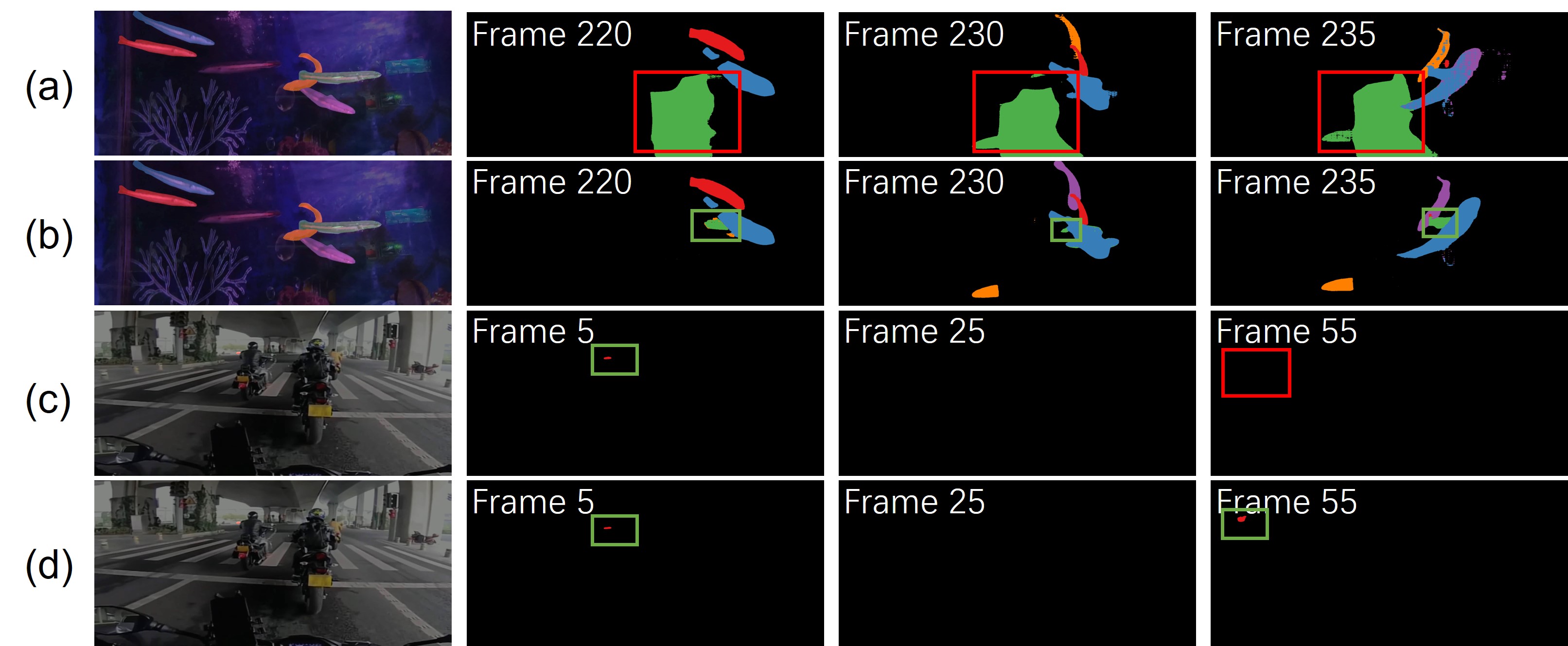}
\caption{Qualitative comparison on video object segmentation. (a), (c) show the results from SAM2, and (b),(d) are drawn from our MoSAM, superior in hard cases including object object disappearance and occlusion. Red boxes suggest the wrong segmentation or object object disappearance, and green boxes indicate accurate segmentation.} 
\label{fig:visualization_2}
\vspace{-2pt}
\end{center}
\end{figure*}

\begin{table}[t]
\renewcommand\arraystretch{1}
\setlength{\tabcolsep}{3.3pt}
\centering
\footnotesize
\caption{Comparison of the MoSAM with previous works on open-vocabulary video instance segmentation task.}
\label{vis-comparison}
\begin{tabular}{lccc|ccc}
\toprule
 \multicolumn{1}{l}{\multirow{2}{*}{Methods}} & \multicolumn{3}{c|}{LV-VIS val} & \multicolumn{3}{c}{LV-VIS test}   \\  
 &  mAP & mAP$_{b}$  & mAP$_{n}$  &  mAP & mAP$_{b}$  & mAP$_{n}$    \\\midrule
Detic-SORT~\cite{sort}  & 12.8 & 21.1 & 6.6   & 9.4  & 15.8 & 4.7    \\
Detic-OWTB~\cite{owtb}  & 14.5 & 22.6 & 8.5   & 11.8 & 19.6 & 6.1    \\ 
Detic-XMem~\cite{XMem}  & 16.3 & 24.1 & 10.6  & 13.1 & 20.5 &  7.7   \\ 
OV2Seg~\cite{ov2seg}    & 21.1 & 27.5 &  16.3 & 16.4 & 23.3 & 11.5   \\ 
OVFormer~\cite{ovformer}& 24.7 & 26.8 &  {23.1} & {19.5} & 23.1 &  {16.7}  \\  
Grounded-SAM2~\cite{grounded-sam2}  & 20.5 & 25.4 & 15.8 & 15.9 & 22.7 &  10.8   \\  
MoSAM   & \textbf{26.4} & \textbf{30.0} &  \textbf{23.7}  & \textbf{20.2} & \textbf{25.8} & \textbf{17.5}   \\  \bottomrule
\end{tabular}
\end{table}

\vspace{-2pt}
\subsection{Comparison with Other Methods on OpenVIS}
\vspace{-2pt}
To validate the generalizability of our approach, we extend MoSAM to the open-vocabulary video instance segmentation (OpenVIS) task, conducting zero-shot transfer evaluations on the LV-VIS validation and test datasets. Specifically, we first maintain a vocabulary list, then utilize the mask predictions from MoSAM to locate the target regions and crop them accordingly. Subsequently, we employ a CLIP~\cite{clip} encoder to obtain visual embeddings, which are then matched for similarity against text embeddings derived from the vocabulary list to predict the categories of each object. Through this method, we achieve OpenVIS and compare our results with previous methods. As reflected in Table~\ref{vis-comparison}, our MoSAM-based model outperforms all others on both the validation and test sets, demonstrating that MoSAM can be easily applied to tasks related to semantic discrimination. It is noteworthy that we follow the evaluation protocol established in previous OpenVIS benchmarks and adopt mAP$_b$, mAP$_n$ to represent base and novel categories, respectively.

\begin{table}[t]
\renewcommand\arraystretch{1}
\setlength{\tabcolsep}{10pt}
\label{tab:ablation}
\centering
\footnotesize
\caption{Ablation study of each proposed strategy in MoSAM, including MGP with sparse motion modeling (SM) and dense motion modeling (DM), and ST-MS with temporal selection (TS) and spatial selection (SS).}
\begin{tabular}{cc|cc|ccc}
\toprule
\multicolumn{2}{c|}{MGP} & \multicolumn{2}{c|}{ST-MS} & \multicolumn{1}{c}{\multirow{2}{*}{$\mathcal{J}\&\mathcal{F}$ }} & \multicolumn{1}{c}{\multirow{2}{*}{$\mathcal{J}$ }} & \multicolumn{1}{c}{\multirow{2}{*}{$\mathcal{F}$ }}  \\
 SM  & DM & TS & SS  &   &  &  \\ \midrule
 &&&                                                     & 80.2 & 75.4 & 84.9 \\
 $\checkmark$ &  &  &                                    & 80.9 & 75.7 & 86.1 \\
 & $\checkmark$ &  &                                     & 81.3 & 76.2 & 86.4 \\
$\checkmark$ & $\checkmark$  &   &                       & 82.1 & 76.6 & 87.5  \\ \midrule

$\checkmark$ &  $\checkmark$ &  $\checkmark$  &          & 84.0 & 79.2 & 88.8 \\
$\checkmark$ &  $\checkmark$ &  &  $\checkmark$          & 82.6 & 77.5 & 87.6 \\ \midrule
$\checkmark$ & $\checkmark$ & $\checkmark$& $\checkmark$ & 84.6 & 79.3 & 89.9 \\ \bottomrule
\end{tabular}
\end{table}

\vspace{-2pt}
\subsection{Visualization} 
\vspace{-2pt}
Figure~\ref{fig:visualization_2} presents a qualitative comparison on LVOS-v1 between our MoSAM with previous approaches~\cite{sam2,yang2024samurai}. For videos with similar nearby objects and long-term occlusions, MoSAM excels in visual object tracking. These challenging scenarios often hinder existing VOS methods from consistently segmenting and tracking objects over time. MoSAM's improvements over the original baseline, visualized through masks, highlight the benefits of incorporating motion prompting and memory selection modules.

\vspace{-2pt}
\subsection{Ablation Study}
\vspace{-2pt}
We conduct detailed ablation studies for the proposed MoSAM using the large size backbone, analyzing the impact of the proposed MGP and ST-MS on LVOS-v1 dataset towards VOS task.

\noindent\textbf{Effectiveness of the MGP} 
We begin with the baseline SAM2 model, incorporating sparse motion modeling to provide sparse point-driven object position predictions, which are put to the model as point prompts. This approach results in improvements of 0.7\%, 0.3\%, and 1.2\% in the $\mathcal{J}\&\mathcal{F}$, $\mathcal{J}$, and $\mathcal{F}$ metrics, respectively, demonstrating that representing and predicting motion using key points can enhance tracking and segmentation capabilities. Alternatively, we augment the baseline with motion represented through dense motion cues, generating box prompts for the model, which yields gains of 1.1\%, 0.8\%, and 1.5\% in three metrics. This indicates that leveraging overall geometric information for motion representation and prediction enables the model to perform better segmentation. Additionally, when both strategies are applied simultaneously, the model exhibits total gains of 1.9\%, 1.2\%, and 2.6\% in three metrics, thereby proving that the complementary property of sparse and dense motion representation and prediction can significantly enhance the model's segmentation performance.

\noindent\textbf{Influence of the ST-MS}
Morever, upon the model with MGP, we combine sparse motion modeling to provide sparse point-driven object position predictions, which are put to the model as point prompts. This approach results in improvements of 0.7\%, 0.3\%, and 1.2\% in the $\mathcal{J}\&\mathcal{F}$, $\mathcal{J}$, and $\mathcal{F}$ metrics, respectively, demonstrating that representing and predicting motion using key points can enhance tracking and segmentation capabilities. Alternatively, we augment the baseline with motion represented through dense motion cues, generating box prompts for the model, which yields gains of 1.1\%, 0.8\%, and 1.5\% in three metrics. This indicates that leveraging overall geometric information for motion representation and prediction enables the model to perform better segmentation. Additionally, when both strategies are applied simultaneously, the model exhibits total gains of 1.9\%, 1.2\%, and 2.6\% in three metrics, thereby proving that the complementary property of sparse and dense motion representation and prediction can significantly enhance the model's segmentation performance.

\vspace{-2pt}
\subsection{Further Analysis}
\vspace{-2pt}

\begin{figure}[t]
\begin{center}
\includegraphics[width=\linewidth]{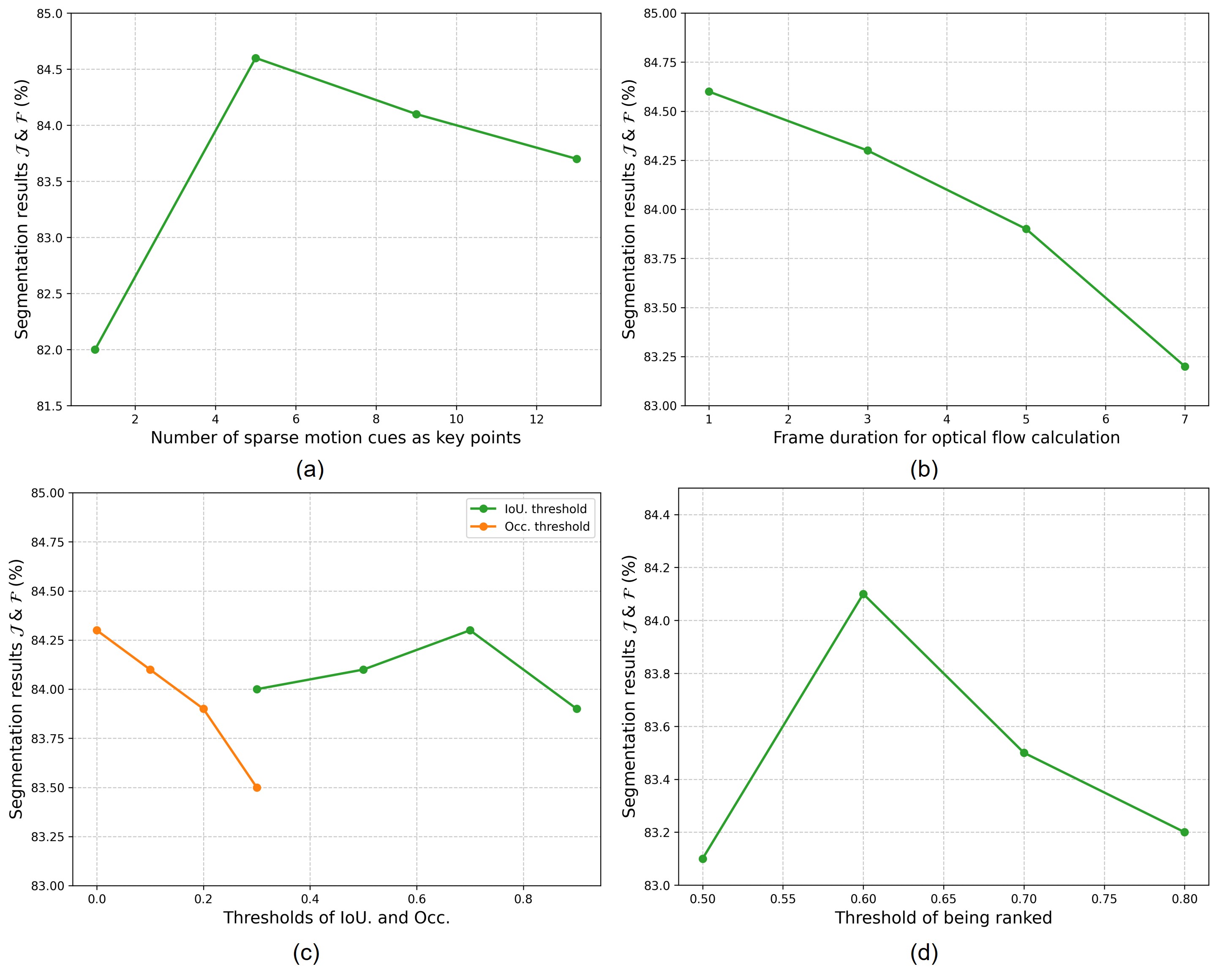}
\caption{Analysis on the hyper-parameters of our MoSAM framework. (a) Number of sparse motion cues as key points; (b) Time interval for optical flow; (c) Thresholds of IoU and Occlusion scores; (d) Threshold of frames to be ranked.} 
\label{fig:analysis}
\vspace{-2pt}
\end{center}
\end{figure}

\noindent\textbf{Number of Sparse Motion Cues}
In MGP, we investigate the impact of the number of key points extracted from sparse motion representation on segmentation performance. We conduct experiments by linearly interpolating to select different quantities of key points in four cardinal directions around the geometric center of the object region. As in Figure~\ref{fig:analysis} (a), the best performance is achieved with five key points; using fewer or more results in slightly reduced performance. This may be attributed to insufficient positional cues with too few key points, while an excess can introduce negative effects from potential incorrect key points.

\noindent\textbf{Time Interval for Optical Flow}
To provide global position information for the object, we calculate optical flow information from the changes in past frames as a dense motion representation. In this process, we analyze the impact of using pairs of frames with different time intervals for optical flow estimation on the final segmentation results. As seen in Figure~\ref{fig:analysis} (b), we find that estimating optical flow using frames spaced one frame apart yields the best results. Longer time intervals result in poorer outcomes, likely due to the sensitivity of global information to long-range motion changes. In contrast, short-term motion variations better characterize and predict future motion trends.

\noindent\textbf{Thresholds of IoU and Occlusion Scores}
In the ST-MS, the thresholds for the IoU and Occlusion (Occ) scores, $\tau_\text{IoU}, \tau_\text{Occ}$ used to filter objects in each frame are crucial, and we conduct a detailed analysis of their effects. For frame-level filtering based on the model's predicted IoU and Occ. scores, we test different thresholds. As suggested in Figure~\ref{fig:analysis} (c), we see that an IoU threshold of 0.7 yields optimal filtering results, while thresholds that are too high or too low result in over-filtering or under-filtering. For Occ. score, the best performance is achieved with a threshold of 0.0, whilst stricter thresholds lead to worse outcomes. Therefore, in our MoSAM, we set the IoU and Occ. thresholds to 0.7 and 0.0, respectively.

\noindent\textbf{Threshold of Frames to Be Ranked}
After the initial threshold screening, the proposed ST-MS strategy performs a secondary removal of frames with an IoU value below a specified threshold. The remaining frames are then ranked based on a combined metric of IoU and Occ. score, selecting the maximum number of frames equal to the proposed quantity to store in the memory bank. To analyze the impact of this threshold on the results, we conducted experiments with different threshold values. As shown in Figure~\ref{fig:analysis} (d), we observe that a threshold of 0.6 yields the best performance, while overly lenient or strict thresholds negatively affected the results. Therefore, we select 0.6 as the threshold for secondary screening.


\vspace{-2pt}
\section{Conclusion}
\vspace{-0.2cm}
\label{sec:conclusion}
In this work, we present MoSAM, which integrates object motion cues and selective memory mechanisms. First, we propose a dual-representation approach that captures both sparse and dense object motion, incorporating them into SAM2 through motion-guided prompts for accurate object tracking. Additionally, we devise a spatial-temporal memory selection mechanism that dynamically filters reliable segmentation results at both pixel and frame levels, ensuring more robust memory features for segmentation. Extensive experiments on video object segmentation and instance segmentation benchmarks demonstrate that MoSAM achieves state-of-the-art performance.

{\small
\bibliographystyle{ieeenat_fullname}
\bibliography{main_arxiv}
}

\end{document}